\documentclass[11pt]{article}
\usepackage{fullpage}
\usepackage[numbers]{natbib}
\usepackage{hyperref}
\usepackage[svgnames]{xcolor}
\usepackage{overpic}
\usepackage{tikz}
\usepackage{rotating}

\hypersetup{colorlinks=true,citecolor=blue,linkcolor=MediumSlateBlue, urlcolor=DarkBlue}

\usepackage{macros}
\usepackage{color}
\usepackage{bbm}
\usepackage{nicefrac}
\usepackage{psfrag}
\usepackage{xifthen} 
\usepackage{xr}
\usepackage{floatrow}
\usepackage{algorithm}
\usepackage{algorithmic}
\usepackage{graphicx}
\usepackage{amsmath}
\usepackage{amsthm}
\usepackage{amssymb}
\usepackage{bbm}
\usepackage{multirow}
\usepackage{capt-of}
\usepackage{sidecap}
\newfloatcommand{capbtabbox}{table}[][\FBwidth]
\sidecaptionvpos{figure}{c}
\newcommand*{\largecdot}{\raisebox{-0.25ex}{\scalebox{1.5}{$\cdot$}}}

\makeatletter
\long\def\@makecaption#1#2{
  \vskip 0.8ex
  \setbox\@tempboxa\hbox{\small {\bf #1:} #2}
  \parindent 1.5em  
  \dimen0=\hsize
  \advance\dimen0 by -3em
  \ifdim \wd\@tempboxa >\dimen0
  \hbox to \hsize{
    \parindent 0em
    \hfil 
    \parbox{\dimen0}{\def\baselinestretch{0.96}\small
      {\bf #1.} #2
    } 
    \hfil}
  \else \hbox to \hsize{\hfil \box\@tempboxa \hfil}
  \fi
}
\makeatother

\long\def\comment#1{}

\begin{document}

\begin{center}
  {\Large{\bf{Stagewise Learning for Sparse Clustering of Discretely-Valued Data}}}

  \vspace*{.3in}
  
  \begin{tabular}{cc}
    Vincent\ Zhao & Steven W.\ Zucker \\
    \href{mailto:yuzhe.zhao@yale.edu}{yuzhe.zhao@yale.edu} &    \href{mailto:zucker@cs.yale.edu}{zucker@cs.yale.edu}
  \end{tabular}
  
  
  
  \vspace*{.2in}
\end{center}

\begin{abstract}  
  The performance of EM in learning mixtures of product distributions often depends on the initialization. This can be problematic in crowdsourcing and other applications, e.g. when a small number of ``experts'' are diluted by a large number of noisy, unreliable participants. We develop a new EM algorithm that is driven by these experts. In a manner that differs from other approaches, we start from a single mixture class. The algorithm then develops the set of ’experts’ in a stagewise fashion based on a mutual information criterion. At each stage EM operates on this subset of the players, effectively regularizing the E rather than the M step. Experiments show that stagewise EM outperforms other initialization techniques for crowdsourcing and neurosciences applications, and can guide a full EM to results comparable to those obtained knowing the exact distribution.
  procedures.
\end{abstract}

%
%
%

\section{Introduction}

We study the model-based sparse clustering problem for discrete data using a mixture model of product distributions \cite{Jain2013,Feldman2005}. This model has application in many fields, including computational neurosciences, crowdsourcing and bioinformatics, and is interesting because it
  differs technically from the problem for continuous data, where the well-known Gaussian mixture model has been applied successfully. 

  A fundamental difficulty is that, in high-dimensional datasets, some features can be noisy, redundant or generally uninformative for clustering, and these can push clustering algorithms toward inappropriate or uninteresting results. If these uninformative or noise data points could be eliminated then, we argue, the results should be much more satisfying. This is precisely our goal: to find an {\em informative set} of data points and to use these to drive the clustering.

\begin{figure}[ht]
    \vskip 0in
    \begin{center}
      \centerline{\includegraphics[width = \columnwidth]{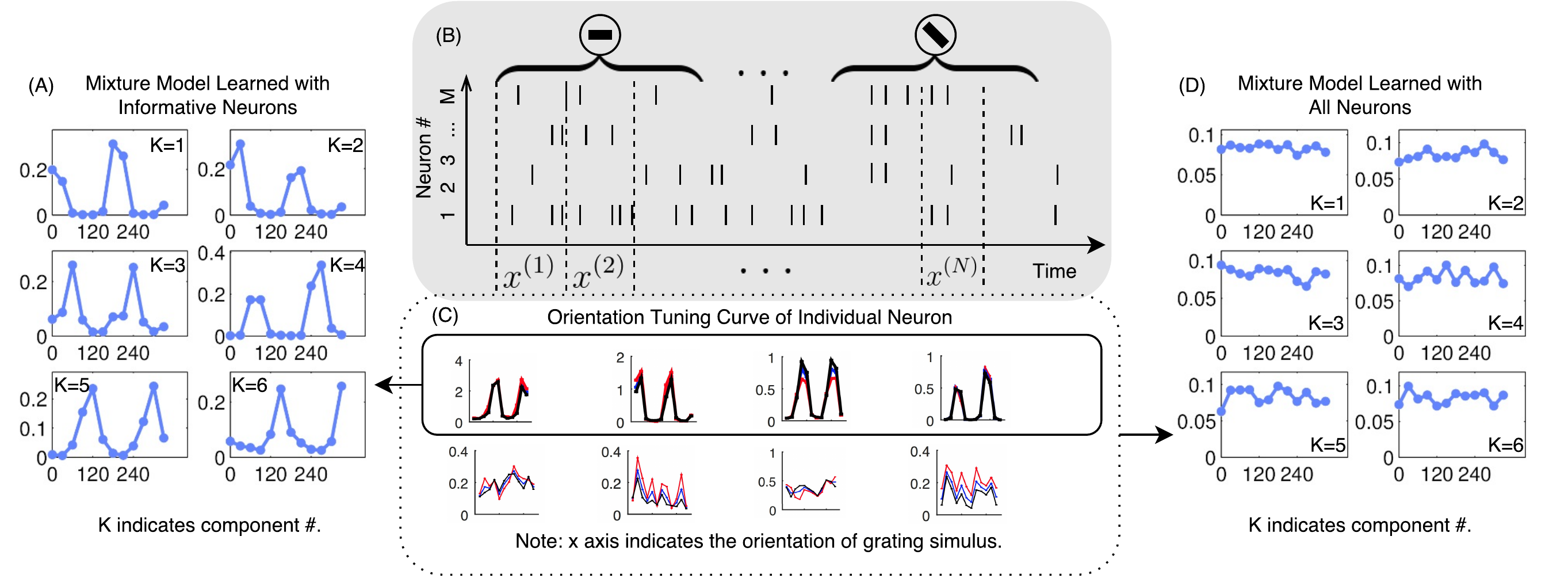}}
      \vskip -0in
      \caption{Bernoulli mixture model learned from multiple neuron spike train data in mouse visual cortex illustrates our goal. Mice view gratings at one of 12 different orientations and an electrode records from multiple units. The gray box illustrates the given data. Some units are selective to the gratings, and some are not (uninformative neurons). If all neurons are used to learn a mixture model, the classes are ill-defined and the orientation tuning curves for each class are uniform (D). (A) shows a mixture model learned by the algorithm developed in this paper, which works by identifying the informative neurons and regularizing EM. In the cartoon these neurons correspond to those that are well tuned, as indicated by the solid box in (C). We emphasize that these tuning curves were not used by our algorithm but were derived from the classes it computed. Data courtesy M. Stryker, University of California at San Francisco. }
      \label{fig:neuron}
    \end{center}
    \vskip -0.2in
  \end{figure}

  We illustrate our goal with a motivating example from neurosciences (Figure \ref{fig:neuron}). Some neurons in mouse visual cortex respond well to certain grating orientations, while others do not respond systematically to gratings. This is called {\em orientation selectivity} \cite{Niell2010}, and we seek to organize neurons into orientation classes according to this activity automatically. The dataset consists of multiple neural spike trains obtained while the mouse viewed gratings with different orientations. The spike trains are converted into binary data indicating the presence (or absence) of an action potential during a short temporal interval. The problem is: given only the spike train data, are we able to group neurons into clusters that correspond to each orientation-tuned class? If we train a mixture model with all neurons using EM, the answer is negative; many of the neurons that are not well-tuned to orientation pollute the results. However, using only the informative neurons (cartooned as those with well-defined tuning curves) we are able to recover the clusters successfully, which underscores our goal to automatically identify those informative and relevant features for discrete data. The algorithms used for this example are developed in this paper, and applied at the end to crowdsourcing data as well.

  A similar problem for continuous data -- the Gaussian mixture model with sparse means -- is better studied. In particular, \cite{Pan2007} proposed an algorithm based on a penalized likelihood function that leads to an EM variant with a regularized M-step, and \cite{Azizyan2013} analyzes learning for a mixture of two isotropic Gaussians in high dimensions under sparse mean separation. As in these papers, we also consider a penalized likelihood function but for a mixture of discrete product distributions. However, we differ from them in that we directly regularize the E-step rather than the M-step. 

  To regularize the E-step, we define an information-theoretic quantity and use it in a novel, stagewise fashion. Our measure is the sum of pairwise conditional mutual information of a certain hybrid distribution, defined below, which turns out to be closely related to maximum likelihood estimation and the EM algorithm.  A similar idea appears in \cite{Witten2010}, who use the Euclidean distance between pairs of data points to regularize a K-means algorithm for sparse clustering. However, in our approach we select which variables to place into the informative set in a stagewise fashion. This stagewise technique is important because many researchers pointed out the drawback of using dimensionality reduction before clustering \cite{Bouveyron2014,Bouveyron2007,Chang1983}. Importantly, as is also explained below, this involves starting with a single class and {\em splitting} it into multiple classes. We stress that our informative set is conceptually different from maximally informative dimensions \cite{Williamson2015}.

  The paper is organized as follows. After briefly introducing the problem setting for learning mixtures of discrete product distributions with sparse structure,  our specific algorithm is presented in \ref{sec:methods}. In \ref{sec:experiments}, we apply our algorithm in crowdsourcing data, to show its range of applicability beyond neurosciences, illustrate our information-theoretic measure and compare it with other state-of-the-art algorithms.

\section{Background}
  \label{sec:background}
  Throughout the paper, we use $[a]$ to denote the integer set $\{1,2,\ldots,a\}$. In a mixture of discrete component distributions (MDPD), it is assumed that each observation $x$ is drawn from a finite mixture distribution $f(X) = \sum_{k=1}^K \omega_k f(X|Y=k;\mu_k)$. $Y\in[K]$ is the latent (non-observable) variable and $\omega_k = f(Y=k)$ denote the mixing weights; they satisfy $\sum_k \omega_k = 1$. We assume $X_i\in[R]$ and $f(X|Y;\mu_k)$ is an $M$-dimensional discrete product distribution that can be factorized as $f(X|Y;\mu_k) = \prod_{i=1}^M f(X_i|Y;\mu_k)$. The conditional distribution, parametrized as $\mu_{kir} = f(X_i = r|Y=k)$ and $\mu_{ki} = [\mu_{ki1}, \ldots, \mu_{kiR]}$, lives on the probability simplex. The set of all parameters is denoted by $\Theta = \{(\omega_k, \mu_k):k\in[K] \}$. Given $N$ observations $[x^{(1)}, \ldots, x^{(N)}]$, the goal of mixture model learning is to maximize the marginal log-likelihood:
  \begin{equation}
  l(\Theta) = \sum_X \hat{f}_0(X) \log \sum_Y f(X,Y;\Theta) =\frac{1}{N} \sum_{n\in[N]} \log \sum_k f(x^{(n)},k;\Theta) .
  \label{eq:loglikelihood}
  \end{equation}
  Since there are latent variables, the marginal log-likelihood is not convex, and EM has been used widely for learning mixture models. EM iteratively updates and optimizes a lower bound of the marginal likelihood function\cite{Minka1998}. The lower bound is obtained by applying Jensen's inequality to the log-likelihood function:
  \begin{equation}
  l(\Theta) \geq \sum_X \hat{f}_0(X) \sum_{Y} q(Y; X, \Theta) \log  \frac{f(X,Y; \Theta)}{q(Y; X, \Theta)} \label{eq:jensen}
  \end{equation}
  where $q(Y; X, \Theta)$ is a distribution over $Y$ that may depend on $X$ or $\Theta$. (We shall work on upper bounds shortly.) Let the current model be parametrized by $\Theta^t$. Then

  \textbf{E-step}: Calculate $f(y|x^{(n)};\Theta^t)$ for $n\in[N] $ and set $q(k; x^{(n)}, \Theta) = f(y=k|x^{(n)};\Theta^t)$.

  \textbf{M-step}: Maximize \eqref{eq:jensen} with regard to $\Theta$.
  \begin{align}
  &\Theta^{t+1} = \arg \max_\Theta Q(\Theta; \Theta^t) \nonumber \\
  &\text{where } Q(\Theta;\Theta^t) = \sum_{X} \hat{f}_0(X) \sum_{Y} f(Y|X; \Theta^t) \log  \frac{f(X ,Y; \Theta)}{f(Y|X; \Theta^t)}
  \label{eq:Q}
  \end{align}
  We study MDPD in a high-dimensional, sparse setting. (The analogous problem for Gaussian mixture models has been studied by \cite{Pan2007,Azizyan2013,Wang2008}.) In this setting, the number of informative variables is much smaller than the dimension $M$. Let $S$ denote the set of informative variables, $|S|<<M$, and let $\bar{S} = \{i\in [M]| i\notin S \}$ be the complementary set. It is intuitive that {\em the uninformative random variables $X_i,\ i\in\bar{S}$ should not be distinguishable across the different mixture components}, i.e. $f_0(X_i|Y=k_1) = f_0(X_i|Y=k_2)$ for $k_1, k_2 \in [K]$. 

  Inspired by \cite{Pan2007}, we consider the following penalized maximum likelihood problem to encourage the sparse structure in the model. 
  \begin{equation}
  \max_\Theta l(\Theta) - \lambda \sum_i ||\sum_k D_{KL}(\bar{\mu}_i||\mu_{ki} )||_0
  \label{eq:penlog}
  \end{equation}
  where $\bar{\mu}_i = \sum_k \omega_k \mu{k_i} $ and $||\largecdot||_0$ is the $l_0$ norm. $D_{KL}(p||q) = \sum p \log (p/q)$ denotes KL-divergence. $D_{KL}(p||q)$ is non-negative and the equality holds if and only if $p=q$. Our penalty encourages sparse structure in the model because $\sum_k D_{KL}(\bar{\mu}_i||\mu_{ki}) = 0$ if and only if $\mu_{ki} = \bar{\mu}_i$ for all $k\in [K]$ which indicates that the conditional probability of the random variable $X_i$ is identical in the different mixture components.

\section{Methods}
  \label{sec:methods}
  For convenience, assume that we can draw an infinite number of samples. To begin,
  \begin{definition}
  Let $f(X,Y;\Theta)$ be a MDPD with parameter $\Theta = {(\omega_k, \mu_k);k\in[K]}$. 

  $X_i$ is informative if and only if $\sum_k D_{KL}(\bar{\mu}_i||\mu_{ki} ) > 0$.
  \end{definition}
  It can be verified that if $X_i$ is uninformative, the marginal distribution $f(X)$ can be factorized as
  \begin{equation}
  f(X) = f(X_{/i})f(X_i).
  \label{eq:factorization}
  \end{equation}
  $S\subseteq [M]$ denotes the index set of informative variables. Due to the factorization \eqref{eq:factorization}, 
  \begin{equation}
  f(Y|X) = f(Y|X_S).
  \label{eq:penestep}
  \end{equation}
  Intuitively, uninformative variables will not affect the posterior distribution -- they provide no information about the underlying latent variable. Now, we analyze the penalized likelihood function \eqref{eq:penlog}. The normal E-step leads to the penalized M-step
  \begin{equation}
  \max_\Theta \sum_X f_0(X) \sum_Y f(Y|X;\Theta^{(t)}) \log \frac{f(X,Y;\Theta)}{f(Y|X;\Theta^{(t)})} - \lambda \sum_i ||\sum_k D_{KL}(\bar{\mu}_i||\mu_{ki} )||_0.
  \label{eq:penalizedlowerbound}
  \end{equation}
  The penalized M-step encourages a sparse update for the model and provides a way to determine $S$. By \eqref{eq:penestep}, it makes the E-step in the next iteration depend only on $X_S$. However, solving \eqref{eq:penalizedlowerbound} is hard, so we seek another way to determine $S$ and, in the process, bypass the penalized M-step by using a regularized E-step, i.e. calculating $f(Y|X_S)$. The following theorem motivates how we select $S$.

  \begin{thm}\footnotemark
  \label{thm:motivation}
  Let $f_0(X)$ be a MDPD from which data are sampled and let $S$ be the informative set of $f_0(X)$. If $f(X_i|Y) = f_0(X_i)$ for $i\in \bar{S}$, then 
  \begin{equation*}
  D_{KL} (f_0(X) || f(X)) = D_{KL} (f_0(X_S) || f(X_S)).
  \end{equation*}
  \end{thm}

  Maximizing the likelihood function is equivalent to minimizing the KL-divergence loss $D_{KL} (f_0(X) || f(X))$, since $D_{KL} (f_0(X) || f(X)) = -H_{f_0}(X) - l(\Theta)$ where $H_{f_0}(X) = \sum_X f_0(X) \log f_0(X) $ is the entropy of $f_0(X)$. The KL-divergence loss can be viewed as a measure of how well the model estimates $f_0(X)$. Theorem \ref{thm:motivation} suggests that, if we have an appropriate model for uninformative features, $S$ could be recovered by solving the following dimensionality reduction problem: find the smallest $S \subseteq [M] $ such that
  \begin{equation*}
  D_{KL} (f_0(X) || f(X)) = D_{KL} (f_0(X_S) || f(X_S)) .
  \end{equation*}
  Although in practice $S$ is unknown, it is easy to find a model $f(X,Y)$ that satisfies the condition in the theorem. Simply pretend that all features are uninformative. Then $f(X,Y)$ is just a one-component MDPD satisfying $f(X_i|Y) = f_0(X_i)$ for $i\in [M]$. We use this idea to initialize our algorithm, which is distinct from the common practice of initializing mixture models with multiple components and random parameters. But two problems arise. First, the one-component model might not be a good one, since it does not capture any high-order interactions. It will have to be split, and the procedure to do this is in Section \ref{sec:algo}. Second, $D_{KL} (f_0(X) || f(X))$ is computationally intractable. Our approach is to find a proper approximation to $D_{KL} (f_0(X) || f(X))$, based on which we can place variables into $S$. The details are as follows.

  \subsection{Conditional Mutual Information Approximates KL-Divergence Loss}
  \label{sec:approx}
  \begin{figure}[ht]
    \floatbox[{\capbeside\thisfloatsetup{capbesideposition={right,center},capbesidewidth=5cm}}]{figure}[\FBwidth]
  {\caption{$U(f_0||\tilde{f}_{S,\Theta^t}) $ is an upper bound of $D_{KL} (f_0 || f_{\Theta^{t+1}}) $ induced by EM. $l(\Theta^*)$ denotes the global maximum of the log-likelihood. And $Q(\Theta,\Theta^t)$ is \eqref{eq:Q} with $f(Y|X;\Theta^t)$ replaced by $f(Y|X_S;\Theta^t)$.}
  \label{fig:bound}}
  {\fbox{\includegraphics[scale=.35]{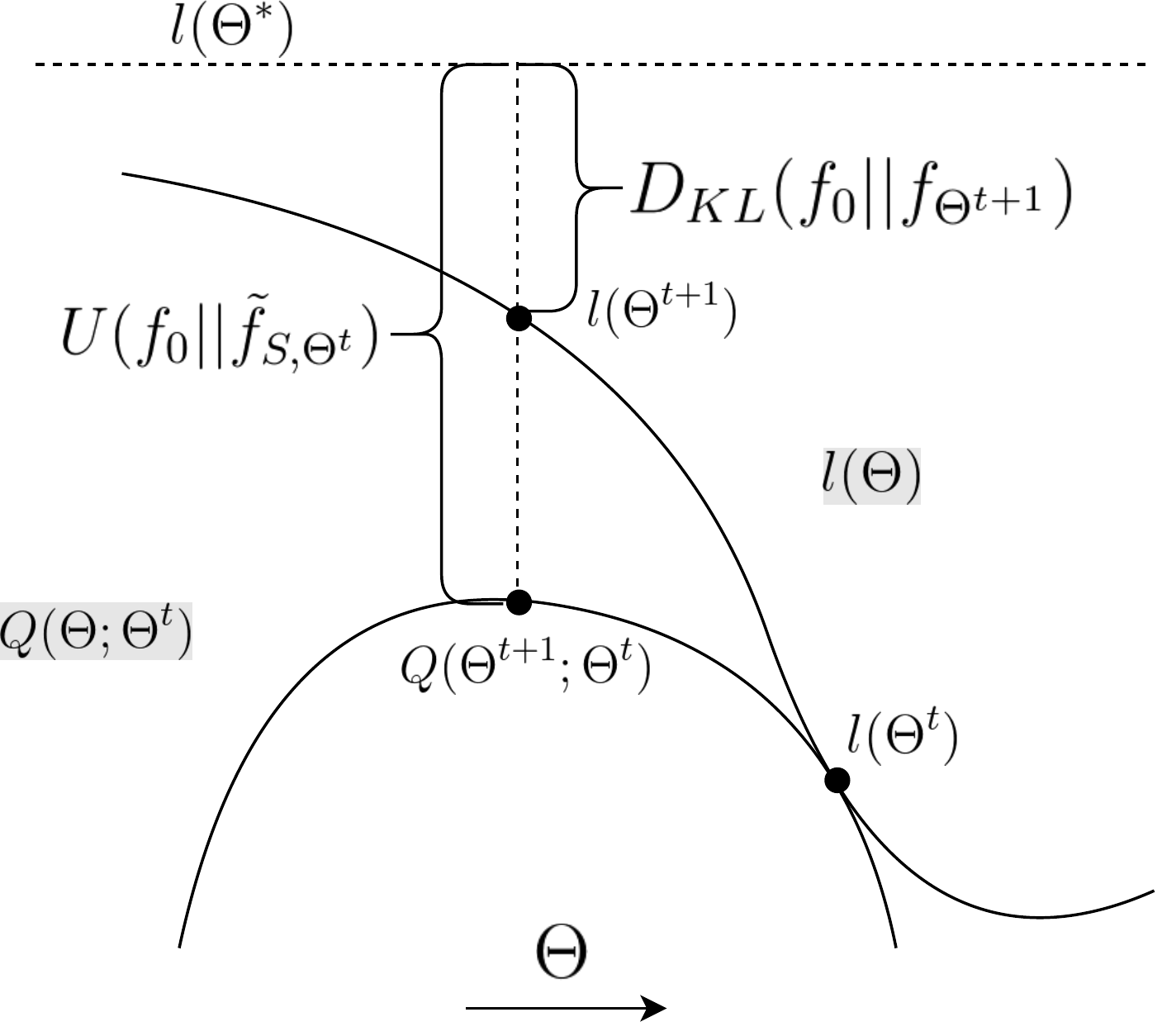}}}
  \end{figure}

  From now on, $D_{KL} (f_0(X) || f(X))$ is referred as $D_{KL} (f_0 || f_{\Theta}) $. We first define a hybrid distribution.

  \begin{definition} The  hybrid distribution is defined as
  \begin{equation}
  \tilde{f}_S (X,Y;\Theta) := f(Y|X_S; \Theta) f_0(X).
  \label{eq:hybrid}
  \end{equation}
  \end{definition}

  The hybrid distribution is a valid probability distribution as it is non-negative and sums to one. By using the hybrid distribution, the following theorem gives an upper bound on $D_{KL} (f_0 || f_{\Theta})$.

  \begin{thm}\footnotemark[\value{footnote}]
  Let $\tilde{f}_S(X,Y;\Theta)$ be the hybrid distribution \eqref{eq:hybrid}, $f(X,Y; \Theta^t)$ be the model distribution at time $t$, and $f(X,Y;\Theta^{t+1})$ be the model distribution after one iteration of EM. Then,
  \begin{align}
  &D_{KL}(f_0||f_{\Theta^{t+1}}) \leq U(f_0||\tilde{f}_{S,\Theta^t}) \nonumber \\
  &\text{where }  U(f_0||\tilde{f}_{S,\Theta^t}) = \sum_{X,Y} \tilde{f}_S(X,Y;\Theta^{t}) \log \frac{\tilde{f}_S(X|Y;\Theta^{t})}{\prod_i \tilde{f}_S(X_i|Y;\Theta^{t})} \nonumber
  \end{align}
  \label{thm:upperbound}
  \end{thm}

  The geometric interpretation of the theorem is provided in Figure \ref{fig:bound}. This theorem is a direct result of Jensen's inequality and the EM algorithm. By information theory, 
  \begin{equation}
  U(f_0||\tilde{f}_{S,\Theta^t}) = \sum_{i\in[M]} H_{\tilde{f}_{S,\Theta^{t}}}(X_i|Y) - H_{\tilde{f}_{S,\Theta^{t}}}(X|Y).
  \label{eq:upperbound}
  \end{equation} 


  The first term in \eqref{eq:upperbound} involves the singleton marginal conditional entropy $H_{\tilde{f}_{S,\Theta^{t}}}(X_i|Y)$ which is computationally tractable. However, because $\tilde{f}(X|Y;\Theta) = \frac{f_0(X)f(Y|X;\Theta)}{\sum_X f_0(X)f(Y|X;\Theta)} $ and $f_0(X)$ cannot be factorized in most cases, the second term $H_{\tilde{f}_{S,\Theta^{t}}}(X|Y)$ is computationally intractable. To tackle the intractability, we further approximate $H_{\tilde{f}_{S,\Theta^{t}}}(X|Y)$ with the Bethe entropy approximation.
  
 Recall, in graphical models,  $X = [X_i]$ are random variables associated with vertices $V$ and $f(X)$ is the joint distribution associated with the graph $G(V,E)$. The Bethe entropy approximation  \cite{Wainwright2008} is defined as 
  \begin{equation*}
  H(X) \approx H_\text{Bethe} = \sum_{i\in V} H(X_i) - \sum_{(s,t)\in E} I(X_s, X_t)
  \end{equation*}
  where $I(X_s, X_t)$ is pairwise mutual information. The Bethe entropy approximation is accurate for acyclic Markov random fields.

 Applying the Bethe entropy approximation to the second term in \eqref{eq:upperbound} yields an approximation to the conditional entropy:

  \begin{align}
  &H_{\tilde{f}_{S,\Theta^{t}}}(X|Y) \approx \sum_i H_{\tilde{f}_{\Theta^t}}\left(X_i|Y)\right) - \sum_{i\neq j} I_{\tilde{f}_{\Theta^t}}(X_i, X_j|Y)
  \label{eq:bethe} \\
  &\text{where } I_{\tilde{f}_{S,\Theta^t}}(X_i, X_j|Y) = \sum_{X,Y} \tilde{f}_S(X,Y;\Theta^t) \log \frac{\tilde{f}_S(X_i,X_j|Y;\Theta^t)}{\tilde{f}_S(X_i|Y;\Theta^t)\tilde{f}_S(X_j|Y;\Theta^t)}. \nonumber 
  \end{align}

  Now, combine \eqref{eq:upperbound} and \eqref{eq:bethe} to approximate an upper bound for the KL-divergence loss:
  
  \begin{equation}
  U(f_0||\tilde{f}_{S,\Theta^t}) \approx \sum_{i\neq j} I_{\tilde{f}_{S,\Theta^t}}(X_i, X_j|Y)
  \label{eq:approx}
  \end{equation}

  The approximation consists of pairwise conditional mutual information. It breaks the curse of dimensionality for KL-divergence loss and the computational complexity of $\sum_{i\neq j} I_{\tilde{f}_{S,\Theta^t}}(X_i, X_j|Y)$ is  $O(KNM^2 R^2)$. It leads to an operational version of Theorem \ref{thm:motivation}:

  \begin{proposition}
    Under the same conditions as in Theorem \ref{thm:motivation}, we have
    \begin{equation}
    \sum_{\substack{i\neq j\\ i,j \in [M]}} I_{\tilde{f}_{S,\Theta^t}}(X_i, X_j|Y) = \sum_{\substack{i\neq j\\ i,j\in S}} I_{\tilde{f}_{S,\Theta^t}}(X_i, X_j|Y)
    \label{eq:prop1}
    \end{equation}
    \label{prop:MIdimred}
  \end{proposition}

 Thus we can recover $S$ in a similar way to that suggested by Theorem \ref{thm:motivation}. Moreover, if the model fits the data perfectly, \eqref{eq:approx} would be zero.

  \begin{proposition}
  If $D_{KL}(f_0||f_{\Theta^{t}}) = 0 $, then $ \sum_{i\neq j} I_{\tilde{f}_{S,\Theta^{t}}}(X_i, X_j|Y) = 0$
  \label{prop:MI is tight}
  \end{proposition}

  In effect from Proposition \ref{prop:MIdimred}, if $I_{\tilde{f}_{S,\Theta}}(X_i, X_j|Y)$ is large for some feature pair $(i,j)$, we can conclude that both $i$ and $j$ are informative. On the other hand, from Proposition \ref{prop:MI is tight}, the model doesn't fit the data well in those dimensions. Therefore, $i$ and $j$ are significant for model learning and should be used to regularize the E-step. This is the key idea that underlies our algorithm.

  \subsection{Algorithm: Stagewise EM}
  \label{sec:algo}
    \begin{algorithm}[ht]
  \caption{stage-wise EM}
  \label{algo:stagewiseEM}
  \begin{algorithmic}
    \STATE {\bfseries INPUT:} $\{x^{(n)} :n\in[N]\}$ and $K_\text{target}$
    \STATE {\bfseries OUTPUT:} $\{\omega_k, \mu_k:k\in[K_\text{target}]\}$
    \STATE {\bfseries Initialization:} $K=1$, $\omega_1 = 1$ and $\{\mu_{1i} = \frac{1}{N} \sum_{n\in [N]} x_i^{(n)}\}$
    \WHILE{Not Converge}
    \STATE Calculate $I_{\tilde{f}_{S,\Theta^t}}(X_i, X_j)|(Y=k)$ for current model.
    \STATE Find the biggest entry $(i,j,k)$ in $I_{\tilde{f}_{S,\Theta^t}}(X_i, X_j)|(Y=k)$.
    \IF{$i\not\in S$ or $j\not\in S$}
    \STATE Add $i$ and $j$ into $S$
    \IF{$K < K_{target}$}
    \STATE Duplicate the $k$-th component and perturb in $i$ $j$ coordinate (explained in context).
    \ENDIF
    \ENDIF
    \STATE Perform regularized E-step and M-step in Theorem \ref{thm:myMstep}
    \ENDWHILE   
  \end{algorithmic}
  \end{algorithm}

  Our main algorithm -- stagewise EM -- is now developed. Following convention (one-hot encoding), let $x_i^{(n)} \in \{0,1\}^R$ be an observation of coordinate $X_i$. The model is initialized as a one-component MDPD such that the conditional distribution of each feature equals the corresponding frequency in the observations. For uninformative features, this initialization is already a good estimate. 

  \begin{thm}\footnotemark[\value{footnote}]
  For finite observations, redefine $\tilde{f}_S (X,Y;\Theta) := f(Y|X_S; \Theta) \hat{f}_0(X)$. The regularized E-step is to calculate $f(Y|X_S=x_S^{(n)};\Theta^t)$ based on current model $f(X,Y;\Theta^t)$ and $S$. The corresponding M-step is given by
  \begin{align*}
  \omega_k^{t+1} &\leftarrow \tilde{f}_S(Y=k;\Theta^t) \\
  \mu_{ki}^{t+1} &\leftarrow \tilde{f}_S(X_i|Y=k;\Theta^t)
  \end{align*}
  \label{thm:myMstep}
  \end{thm}

  Thus stagewise EM iteratively performs a regularized E-step followed by a corresponding M-step. But, to regularize the E-step, the informative set $S$ has to be obtained explicitly, which we do in an interlaced fashion together with the EM iterations. Specifically, since at least one EM iteration is needed for each update of $S$, the algorithm works conservatively and attempts to update $S$ after each iteration of EM. 

 We now develop the update for informative set $S$.  By a standard result in information theory, 
  \begin{equation}
  I_{\tilde{f}_{S,\Theta^t}}(X_i, X_j|Y) = \sum_k \tilde{f}_S(Y=k;\Theta^t) I_{\tilde{f}_{S,\Theta^t}}(X_i, X_j)|(Y=k)
  \end{equation}
  where $I_{\tilde{f}_{S,\Theta^t}}(X_i, X_j)|(Y=k) = \sum_{X_i,X_j} \tilde{f}_S(X_i,X_j|Y=k;\Theta^t) \log \frac{\tilde{f}_S(X_i,X_j|Y=k;\Theta^t)}{\tilde{f}_S(X_i|Y;\Theta^t)\tilde{f}_S(X_j|Y=k;\Theta^t)}$. $S$ is updated by picking the biggest triplet $(i,j,k)$ in $I_{\tilde{f}_{S,\Theta^t}}(X_i, X_j)|(Y=k)$ and adding the related indices, $i$ and $j$, into $S$ (if they are not already in $S$). The stagewise update is a strong regularization on EM, as it enforces EM on the features that are informative and have not been fitted well in the current model. We use the word ``stagewise'' because a similar idea has been applied to regression \cite{Efron2004,Hastie2009}.

 An important detail remains. Since the algorithm is initialized as a one-component model, or during iterations, we may need to increase the number of components. We do this by splitting one, and again act conservatively:
First find the largest triplet $(i,j,k)$, duplicate the $k$-th component, and add it  into the mixture model: $\mu_{\text{new} i} \leftarrow \mu_{ki}^{(t)} $ for $i\in[M] $ and $\omega_{new}, \omega_k^{t+1} \leftarrow 0.5\omega_k^t $.

  \begin{thm}\footnotemark[\value{footnote}]
  Let $f(X,Y;\Theta^\text{old}) $ be the model before duplication and $f(X,Y;\Theta^\text{new}) $ the one after duplication. It can be shown that
  \begin{equation*}
  \sum_{i\neq j} I_{\tilde{f}_{S,\Theta^\text{old}}}(X_i, X_j|Y) = \sum_{i\neq j} I_{\tilde{f}_{S,\Theta^\text{new}}}(X_i, X_j|Y)
  \end{equation*}\label{thm:duplication}
  \end{thm}
  \footnotetext{See proofs in the supplement}

  Intuitively, since the duplication does not alter the marginal distribution $f(X;\Theta)$, the KL-divergence loss remains unchanged. Theorem \ref{thm:duplication} indicates that $\sum_{i\neq j} I_{\tilde{f}_{S,\Theta}}(X_i, X_j|Y)$ also remains the same. To break the symmetry between the $k$-th component and the new component, we freeze all parameters $\Theta$ in $\sum_{i\neq j} I_{\tilde{f}_{S,\Theta}}(X_i, X_j|Y)$ except for $\mu_{ki}$, $\mu_{kj}$, $\mu_{\text{new}i} $, and $\mu_{\text{new}j} $ and perturb the model with regard to the free parameters. Due to the symmetry, it can be shown that $\Theta^\text{new}$ (parameters after duplication) is a saddle point to the restricted function. Therefore, we calculate the Hessian of $\sum_{i\neq j} I_{\tilde{f}_{S,\Theta}}(X_i, X_j|Y)$ with regard to the free parameters and perturb in the direction of the eigenvector with the most negative eigenvalue of the Hessian.

\section{Empirical Studies}
 \label{sec:experiments}
  Following \cite{Dawid1979}, model-based learning in crowdsourcing can be viewed as a special case of MDPD. Now $X_i\in[R]$ is the label given by the $i$-th worker ($i\in [M]$) to an item with true label denoted by $Y\in[K]$; this requires $R=K$. The goal is to estimate the true label for each question and to assess the individual workers' performances. All the examples below are $\alpha$-sparse crowdsourcing, in which only $\lceil \alpha M \rceil$ workers give the true label with some probability; the other workers give random labels with unknown probability. As we show, stagewise EM performs well against the state-of-the-art crowdsourcing algorithms.

  We first study the behavior of the informative set $S$ using a simulation of 0.3-sparse MDPD $f_0(X)$ with 3 components ($K=3$). 100 workers provide labels to 1000 items with the 30 informative (``expert'') workers enjoying decreasing capabilities: the first worker provides the true label with probability 0.7 and the 30th worker with probability 0.45. The rest of the workers are random.
  \begin{figure}[ht]
  \vskip 0.2in
  \begin{center}
    \centerline{\includegraphics[width = \columnwidth]{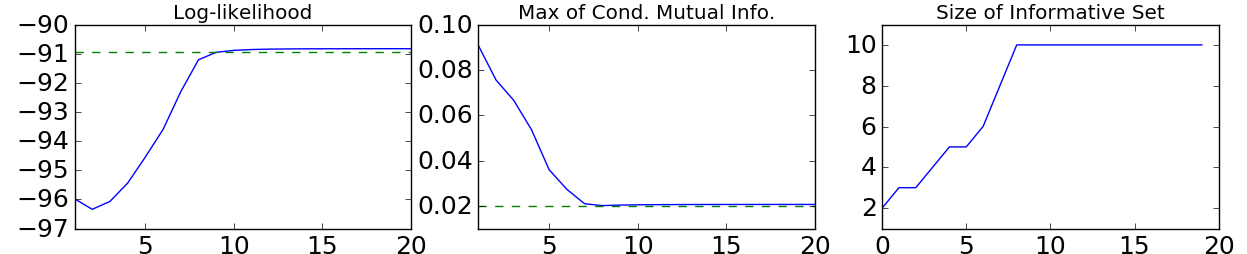}}
    \caption{The performance of stagewise EM on $0.3$-sparse MDPD: from the left to the right are log-likelihood, max norm of conditional mutual information, and the size of the informative set $S$ against the number of iterations. Dashed lines are benchmarks obtained from the underlying true distribution.}
    \label{fig:exp1}
  \end{center}
  \vskip -0.2in
  \end{figure}

  \begin{figure}[ht]
  \vskip -0.2in
  \begin{center}
    \centerline{\includegraphics[width = \columnwidth]{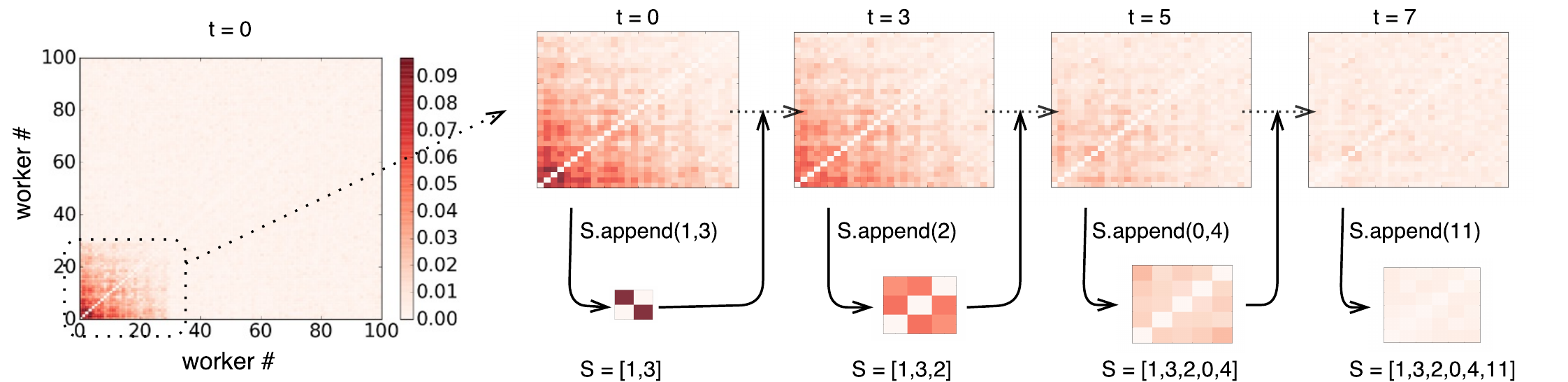}}
    \caption{Illustration of how $S$ evolves and regularizes EM. The diagonal entries are of no interests and therefore eliminated. The leftmost panel shows the conditional mutual information $I_{\hat{f}_0}(X_i,X_j)$ at $t=0$.  The next four top panels show the $I_{\tilde{f}_{S,\Theta^t}}(X_i,X_j|Y)$ for the first 30 workers at iteration $t=0,3,5,7$, while the bottom panels present the mutual information among workers in $S$. For each iteration, the informative set $S$ regularizes the E-step. }
    \label{fig:infoset}
  \end{center}
  \vskip -0.2in
\end{figure}

 We perform 20 iterations of stagewise EM  (Figure
 \ref{fig:exp1}). The benchmark log-likelihood is given by $\sum_X
 \hat{f}_0(X) \log f_0(X) = \frac{1}{100}\sum_{n=1}^{100} \log
 f_0(x^{(n)})$, and the benchmark conditional mutual information
 (middle panel) is also obtained with the true distribution and
 training data. According to Proposition \ref{prop:MI is tight}, the
 max norm of the conditional mutual information evolves toward 0. The
 algorithm converges within 10 iterations and the size of $|S| <
 10$. A more detailed look at the  mutual information criterion
 \eqref{eq:approx} for dimensionality reduction is illustrated in
 Figure \ref{fig:infoset}. By construction, the workers' capabilities
 decay as the index increases. Note how stagewise EM rapidly
 identifies the top 5 most informative workers. The full $S$ for this
 task is (in order) $[1, 3, 2, 0, 4, 11, 6, 15, 48, 77]$. The first 8
 in $S$ are all top 15 workers and, from Figure \ref{fig:exp1}, the
 algorithm almost converges at that point; the later members of
 $S$ are not important. This further suggests that the algorithm seems
 able to ``decide'' how much information is needed to learn the
 model; although 30 informative workers exist, practically  less than 10 are needed for a good estimation.

\begin{figure}
\vskip -0in
  \begin{floatrow}
  \ffigbox{%
  \fbox{\includegraphics[width = \columnwidth]{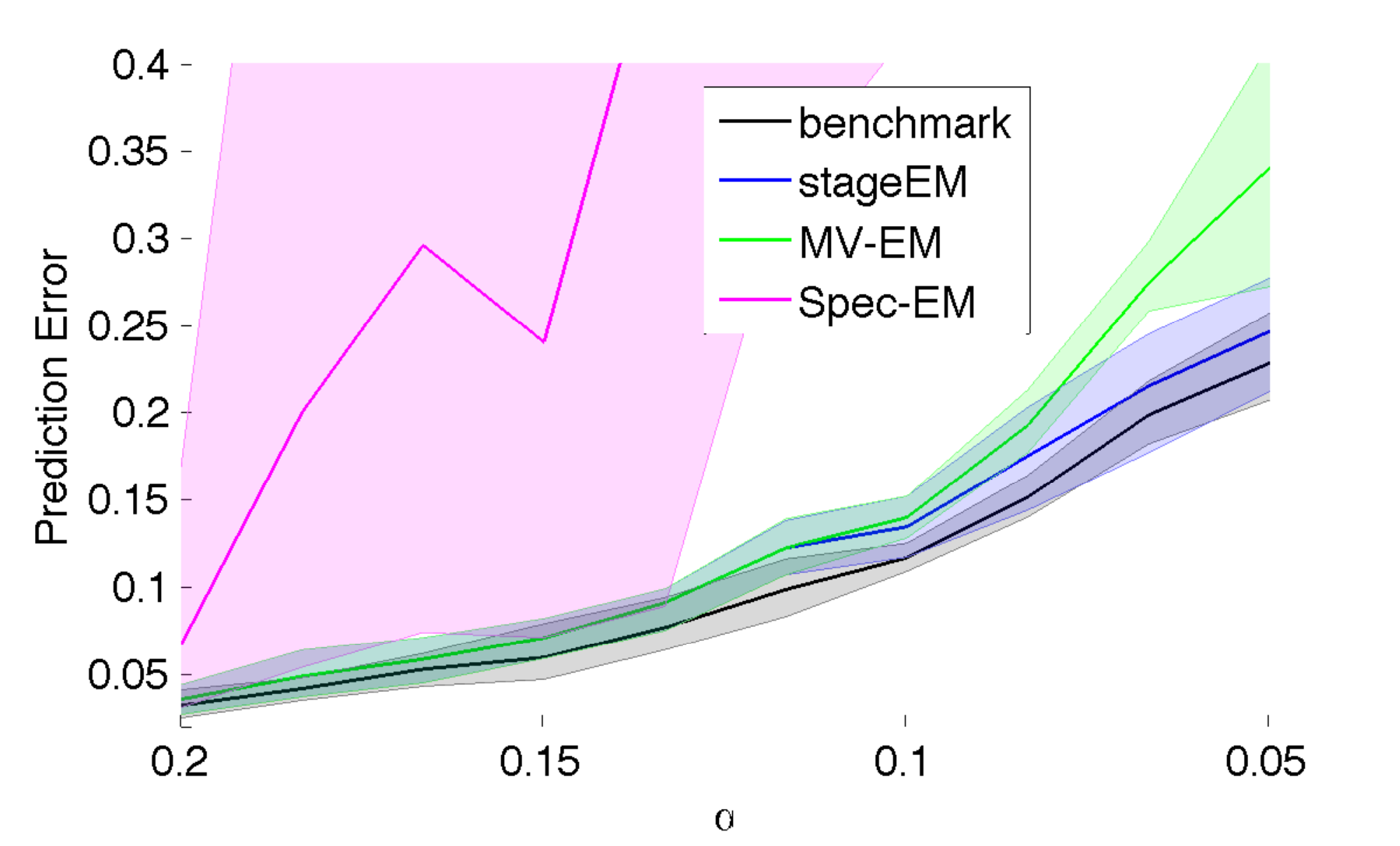}}
  }{%
    \caption{Prediction performance on 3-label $\alpha$-sparse crowdsourcing model: For each $\alpha$, experiments are repeated 10 times. The shaded error bar shows the best and the worst performance.}\label{fig:synthetic}
  }

  \capbtabbox{%
  \begin{tabular}{ccccr}
          \hline
          \multicolumn{2}{p{2cm}}{Algorithm} & Bird & Dog  \\
          & &(2 labels) & (4 labels)\\
          \hline\hline
          \multicolumn{2}{p{2cm}}{StageEM-refine}    &\textbf{10.19} & 16.73 \\ \hline
          \multicolumn{2}{p{2cm}}{StageEM}        &12.04  &20.69  \\ 
          \multicolumn{2}{p{2cm}}{($|S|/M$)}        &(11/39) &(14/52)   \\ \hline
          \multirow{3}{*}{Spec-EM}   &Low      &11.11  &16.98  \\
                &Average    &11.57  &22.19  \\
                & High      &12.04  &31.85  \\ \hline
          \multicolumn{2}{p{2cm}}{MV-EM\footnote{Refer to the results reported in \cite{Zhang2014}}}      & 11.11 & \textbf{16.66} \\
          \hline\hline
        \end{tabular}
  }{%
    \caption{Prediction error $(\%)$ on real datasets. $|S|$ is the size of informative set, while $M$ is the total number of workers.\label{tb:realdata}}%
  }
  \end{floatrow}
\vskip -0.2in
\end{figure}

  We now study the prediction performance by comparing
  our algorithm (stage-EM) against the spectral algorithm (Spec-EM)
  \cite{Zhang2014} and the majority-vote-initialized EM (MV-EM) commonly used in practice. We start with synthetic data
  for the $\alpha$-sparse crowdsourcing problem with 100 workers, 1000 samples, and 3 labels. The first $100\alpha$ workers are
  informative giving correct labels with probability 0.6; the rest are uninformative workers
  giving labels at random with unknown probability. We vary $\alpha\in [0.05,0.2]$ and,
  for each $\alpha$, the experiment is repeated for 10 times. In Figure
  \ref{fig:synthetic}, we show prediction performances achieved by
  different algorithms. The benchmark score is the prediction error by
  the true model. Spec-EM does not work in this sparse setting, MV-EM is
  able to keep up with our algorithm until $\alpha$ becomes small, while
  stage-EM stays close to the benchmark for all $\alpha$. Our algorithm
  consistently outperforms the other methods in this sparse setting.

Finally, we turn to real datasets (Table \ref{tb:realdata}), even though they are not sparse.
The bluebird dataset \cite{Welinder2010} is a
binary labeling task containing 108 items, 39 workers and 4,212
observed labels. The dog dataset \cite{Zhou2012} contains 4 different
dog breed labels from ImageNet. Since these datasets are incomplete,
we add a new ``missing label'' which indicates that the worker does
not label this item. The probability of not giving a label is assumed
to be independent of the true label. It is estimated from the data for
each worker and then frozen (not trainable) during model fitting.
Since these datasets are not sparse, we run regular EM on the complete dataset after model fitting to leverage all the information (StageEM-refine). 
As shown in Table \ref{tb:realdata}, it is still comparable with MV-EM and surpasses
Spec-EM. Importantly, stagewise EM has decent prediction performance
using only about 1/3 of the workers available in both datasets.

\section{Conclusion}
We developed a stagewise EM algorithm for sparse clustering of discretely-valued data.
The key insight is that uninformative features should have uniform probability of belonging
to any mixture class. This led to an informative set of features via a mutual information criterion and a practical algorithm by approximating it with Bethe entropy. The result performed well for neurosciences and crowdsourcing datasets.

\section{Acknowledgements}
We thank M. Stryker and M. Dadarat for the providing the neural data, and Harrison Huibin Zhou and Sahand N. Negahban for suggestions. We also thank Junjiajia Long for dicussions. Research supported by the Simons Foundation, the Paul Allen Family Foundation, and China Scholarship Council.



\bibliography{cite.bib}
\bibliographystyle{plain}

\newpage
\section*{Supplements}
\subsection*{Proof: Theorem \ref{thm:motivation}}
Since $S$ is the informative set, by \eqref{eq:factorization}, 
\begin{equation*}
  f_0(X) = f_0(X_S) \prod_{i\in\bar{S}}f_0(X_i). 
\end{equation*}
In the theorem, we set $f(X_i|Y) = f_0(X_i)$ for $i\in \bar{S}$. Hence, we also have 
\begin{align*}
  f(X) &= f(X_S) \prod_{i\in\bar{S}}f(X_i) \\
  &= f(X_S) \prod_{i\in\bar{S}}f_0(X_i). 
\end{align*}
For KL-divergence,
\begin{align*}
D_{KL} (f_0(X) || f(X)) &= \sum_X f_0(X) \log\frac{f_0(X)}{f(X)} \\
&= \sum_X f_0(X) \log\frac{f_0(X_S) \prod_{i\in\bar{S}}f_0(X_i)}{f(X_S) \prod_{i\in\bar{S}}f(X_i)} \\
&= \sum_X f_0(X) \log\frac{f_0(X_S)}{f(X_S)} \\
&= D_{KL} (f_0(X_S) || f(X_S))
\end{align*}

\subsection*{Proof: Theorem \ref{thm:upperbound}}
\begin{equation*}
D_{KL}(f_0||f_{\Theta^{t+1}}) = -H_{f_0}(X) - l(\Theta^{t+1})
\end{equation*}
By the standard result from EM, $l(\Theta^{t+1})$ is lower bounded
\begin{equation*}
l(\Theta^{t+1}) \geq \max_\Theta Q(\Theta; \Theta^t)
\end{equation*}
where $Q(\Theta;\Theta^t) = \sum_{X} f_0(X) \sum_{Y} f(Y|X_S; \Theta^t) \log  \frac{f(X ,Y; \Theta)}{f(Y|X; \Theta^t)}$. Let $\tilde{f}_S(X,Y;\Theta)$ be defined as \eqref{eq:hybrid} and by Theorem \ref{thm:myMstep},
\begin{equation*}
\max_\Theta Q(\Theta;\Theta^t) = \sum_{X,Y} \tilde{f}_S(X,Y; \Theta^t) \log  \frac{\prod_{i\in[M]} \tilde{f}_S(X_i|Y;\Theta^t)\tilde{f}_S(Y;\Theta^t)}{f(Y|X; \Theta^t)}
\end{equation*}
Using the lower bound of $l(\Theta)$ to upper bound $D_{KL}(f_0||f_{\Theta^{t+1}})$ gives the desired result.

\subsection*{Proof: Theorem \ref{thm:myMstep}}
In finite observation case, after a regularized E-step, M-step is to
\begin{equation*}
\max_\Theta \frac{1}{N} \sum_n \sum_{k\in[K]} f(y=k|x^{(n)};\Theta^t) \log f(x^{(n)},y=k;\Theta) 
\end{equation*}
By the factorization of MDPD, it can be written as 
\begin{equation*}
\max_\Theta \frac{1}{N} \sum_n \sum_{k\in[K]} f(y=k|x^{(n)};\Theta^t) \left( \sum_{i\in[M]}\log f(x_i^{(n)}|k;\Theta) + \log f(y=k;\Theta) \right).
\end{equation*}
By the parameterization of MDPD, the discrete distribution $f(x_i|k)$ is represented by $\mu_{ki} = [\mu_{ki1}, \ldots, \mu_{kiR}]^T $ and $f(y)$ by $\omega = [\omega_{k1}, \ldots,\omega_{k1}] $, which satisfying
\begin{align*}
\sum_r \mu_{kir} &= 1 &\text{and}& &\sum_k \omega_k = 1.
\end{align*}
Therefore, we can maximize over different $\mu_{ki}$ and $\omega$ separately. This constrained optimization problem is solved by applying Lagrangian multiplier, which gives 
\begin{align*}
\mu_{kir} &\leftarrow \frac{1/N\sum_n f(y=k|x^{(n)};\Theta^t) \mathbbm{1}_{x_i^{(n)} = r}}{\sum_r 1/N\sum_n f(y=k|x^{(n)};\Theta^t) \mathbbm{1}_{x_i^{(n)} = r}} \\
\omega_k &\leftarrow \frac{1/N\sum_n f(y=k|x^{(n)};\Theta^t)}{\sum_k 1/N\sum_n f(y=k|x^{(n)};\Theta^t)}.
\end{align*}
$\mathbbm{1}$ is indicator function. We define $\tilde{f}_S (X,Y;\Theta) := f(Y|X_S; \Theta) \hat{f}_0(X)$, therefore the probability for any instance $(x,y)$ is 
\begin{equation*}
\tilde{f}_S (X=x,Y=y;\Theta) = \frac{1}{N} \sum_n f(y|x^{(n)}_s;\Theta) \mathbbm{1}_{x^{(n)}=x}.
\end{equation*}
And we can reformulate the results as 
\begin{align*}
  \omega_k^{t+1} &\leftarrow \tilde{f}_{S,\Theta^t}(Y=k) \\
  \mu_{kir}^{t+1} &\leftarrow \tilde{f}_{S,\Theta^t}(X_i=r|Y=k)
\end{align*}

\subsection*{Proof: Theorem \ref{thm:duplication}}
It is enough to prove that for all pair $(i,j)$, 
\begin{equation}
  I_{\tilde{f}_{S,\Theta^\text{old}}}(X_i, X_j|Y) = I_{\tilde{f}_{S,\Theta^\text{new}}}(X_i, X_j|Y).
  \label{eq:duplication}
\end{equation}
The conditional mutual information can be decomposed as 
\begin{equation*}
  I_{\tilde{f}_{S,\Theta}}(X_i, X_j|Y) = \sum_k \tilde{f}_S(Y=k;\Theta) I_{\tilde{f}_{S,\Theta}}(X_i, X_j)|(Y=k).
\end{equation*}
Let the component $k_2$ be the duplication of the component $k_1$. We notice that 
\begin{equation*}
  \tilde{f}_S(Y=k;\Theta^{new}) =
  \begin{cases}
  \tilde{f}_S(Y=k;\Theta^\text{old}), &\text{ if }k \neq k_1 \text{ or }k_2\\
  \frac{1}{2}\tilde{f}_S(Y=k_1;\Theta^\text{old}), &\text{ if }k = k_1 \text{ or }k_2
  \end{cases}.
\end{equation*}
Moreover, duplication does not change the conditional distribution within each component, so we have
\begin{equation*}
  I_{\tilde{f}_{S,\Theta^\text{new}}}(X_i, X_j)|(Y=k) =
  \begin{cases}
    I_{\tilde{f}_{S,\Theta^\text{old}}}(X_i, X_j)|(Y=k), &\text{ if }k \neq k_1 \text{ or }k_2\\
    I_{\tilde{f}_{S,\Theta^\text{old}}}(X_i, X_j)|(Y=k_1), &\text{ if }k = k_1 \text{ or }k_2
  \end{cases}.
\end{equation*}
Therefore, we know that \eqref{eq:duplication} holds.

\subsection*{Proof: Proposition \ref{prop:MIdimred}}
The proposition follows from the fact that if $i\in\bar{S}$, then
\begin{equation*}
  \tilde{f}_{S,\Theta}(X,Y) = f(Y|X_S;\Theta)f_0(X_{/i}) f_0(X_i).
\end{equation*}
The only term containing $X_i$ is $f_0(X_i)$. Therefore, $X_i$ is independent of other features in the hybrid distribution, which leads to the proposition.

\subsection*{Proof: Proposition \ref{prop:MI is tight}}
If $D_{KL}(f_0||f_{\Theta^{t}}) = 0 $, then $f(X;\Theta^t) = f_0(X)$. And the hybrid distribution becomes
\begin{equation*}
  \tilde{f}_{S,\Theta^t}(X,Y) = f(Y|X_S;\Theta^t)f(X;\Theta^t)
\end{equation*}
By Proposition \ref{prop:MIdimred}, it is enough to show that $I_{\tilde{f}_{S,\Theta^t}}(X_i, X_j|Y) = 0$ for $\{(i,j)| i,j\in S\text{ and }i\neq j \}$. This is true because $X_i$ and $X_j$ is independent conditional on $Y$.


\end{document}